\documentclass{IEEEtran}
\usepackage{graphicx} 
\usepackage{biblatex}
\usepackage{amsmath}
\usepackage{amssymb}
\usepackage{array}
\usepackage{color}
\usepackage{booktabs}
\usepackage{array, caption, threeparttable}
\usepackage[font=small,labelfont=bf,labelsep=none]{caption}

\newtheorem{Proposition}{\textbf{Proposition}}
\newtheorem{Lemma}[Proposition]{\textbf{Lemma}}
\newtheorem{Corollary}[Proposition]{\textbf{Corollary}}
\usepackage{stfloats}
\usepackage{subcaption}
\usepackage{cuted}
\usepackage{algorithmic}
\usepackage{algorithm}

\makeatletter
\newenvironment{breakablealgorithm}
{
		\begin{center}
			\refstepcounter{algorithm}
			\hrule height.8pt depth0pt \kern2pt
			\renewcommand{\caption}[2][\relax]{
				{\raggedright\textbf{\ALG@name~\thealgorithm} ##2\par}%
				\ifx\relax##1\relax 
				\addcontentsline{loa}{algorithm}{\protect\numberline{\thealgorithm}##2}%
				\else 
				\addcontentsline{loa}{algorithm}{\protect\numberline{\thealgorithm}##1}%
				\fi
				\kern2pt\hrule\kern2pt
			}
		}{
		\kern2pt\hrule\relax
	\end{center}
}

\title{Learning-Enhanced Observer for Linear Time-Invariant Systems with Parametric Uncertainty}
\author{Hao Shu\thanks{Hao Shu: Hao$\_$B$\_$Shu@163.com}\IEEEauthorrefmark{1}
\\
\IEEEauthorblockA{\IEEEauthorrefmark{1} Sun-Yat-Sen University, Shenzhen, China.}
}
\date{}

\begin{document}

\maketitle

\begin{abstract}
This work introduces a learning-enhanced observer (LEO) for linear time-invariant systems with uncertain dynamics. Rather than relying solely on nominal models, the proposed framework treats the system matrices as optimizable variables and refines them through gradient-based minimization of a steady-state output discrepancy loss. The resulting data-informed surrogate model enables the construction of an improved observer that effectively compensates for moderate parameter uncertainty while preserving the structure of classical designs. Extensive Monte Carlo studies across diverse system dimensions show systematic and statistically significant reductions, typically exceeding 15\%, in normalized estimation error for both open-loop and Luenberger observers. These results demonstrate that modern learning mechanisms can serve as a powerful complement to traditional observer design, yielding more accurate and robust state estimation in uncertain systems. Codes are available at \url{https://github.com/Hao-B-Shu/LTI_LEO}.
\end{abstract}


\section{Introduction}

State estimation is a foundational component of modern control theory, forming the backbone of feedback regulation, fault detection, and system-level monitoring. When the full system state is not directly measurable, an observer must be designed to reconstruct the hidden state trajectory from available input–output data. For linear time-invariant (LTI) systems, the classical Luenberger observer~\cite{Luenberger1964Observing} remains one of the most widely adopted solutions due to its simple structure and guaranteed asymptotic convergence under the standard observability condition. When the system matrices are known exactly, the observer gain can be readily selected to ensure stable estimation error dynamics.

In practice, however, exact models of real-world systems are seldom available. Parametric discrepancies arise naturally from modeling simplifications, unmodeled dynamics, environmental variations, and sensor noise. These uncertainties lead to nominal system matrices that differ from the true ones, and even small deviations in $(A,B,C)$ may deteriorate the convergence of a Luenberger observer. In particular, biased or unstable estimation trajectories can occur when the nominal parameters drift away from the actual system dynamics.

Several strategies have been proposed to address this challenge. Stochastic methods, most notably the Kalman filter~\cite{Kalman1960A}, optimally fuse model-based predictions with sensor measurements. Nevertheless, these techniques presuppose accurate statistical information about process and measurement noise and remain sensitive to model mismatch. A common alternative is to identify the system parameters offline and subsequently design a nominal observer. Yet system identification generally requires access to full or partial state information~\cite{Farison1967Identification,Chu1994Simultaneous}, and even when successful, the identified model may only reproduce input–output behavior rather than reproduce the internal dynamics faithfully~\cite{Nagumo1967A,Mohamed2023The}. Such discrepancies can pose intrinsic limitations for observer design. Adaptive observers constitute another classical line of work, where both the state and unknown parameters are updated online through suitable adaptation laws. These methods, however, typically rely on known parameterization structures~\cite{Carroll1973An,Narendra1976Stable,Kreisselmeier1977Adaptive,Nuyan1979Minimal}, specific uncertainty models~\cite{Katiyar2023Initial}, or prespecified noise characteristics~\cite{Romero2025An}. These assumptions are often restrictive in scenarios where only coarse or noisy system estimates are available.

Motivated by the need for an observer design method that remains effective under modest but non-negligible parameter uncertainty, we propose a learning-enhanced observer(LEO) framework. The key idea is conceptually simple: treat the nominal matrices $(A,B,C)$ as optimization variables, compute a steady-state discrepancy measure between the observer-generated outputs and the true outputs, and iteratively refine the parameter estimates through gradient-based learning. Once the parameters converge, an improved Luenberger observer is reconstructed from the optimized model.

This learning-enhanced procedure leverages the expressive flexibility of gradient-based optimization while maintaining the structural clarity and theoretical interpretability of the classical Luenberger observer. Extensive numerical experiments demonstrate that the proposed method consistently reduces estimation error, often by more than 20\%, across a wide range of system dimensions.

The paper is organized as follows: 
Section~\ref{sec:formular} formulates the problem. Section~\ref{sec:method} presents the proposed learning-based optimization framework. Section~\ref{sec:Experiment} provides comprehensive numerical validation, and Section~\ref{sec:conclusion} concludes the paper.

\section{\label{sec:formular}Problem Formulation}

Consider a discrete-time linear time-invariant (LTI) system described by
\begin{equation}
\label{eq:discreteLTI}
    x_{k+1} = A x_{k} + B u_{k}, \ y_{k} = C x_{k}
\end{equation}
Although the theoretical development extends naturally to continuous-time systems,
\begin{equation}
\label{eq:continuousLTI}
\dot{x}(t) = A x(t) + B u(t),\ y(t) = C x(t)
\end{equation}
Our focus will remain on the discrete-time model~\eqref{eq:discreteLTI} for convenience, where $x_{k}\in \mathbb{R}^{n}$ denote the system states, $y_{k} \in \mathbb{R}^{q}$ are the outputs, $u_{k} \in \mathbb{R}^{p}$ is the control input, and $A \in \mathbb{R}^{n\times n}, B \in \mathbb{R}^{n\times p}, C \in \mathbb{R}^{q\times n}$ are the system matrices, which are assumed to be time-invariant. The initial state $x_0$ is unknown.

In realistic scenarios, the true system is affected by process and measurement noise. Therefore, a more realistic model is
\begin{equation}
\label{eq:noisyTrueSystem}
x_{k+1} = A_{\mathrm{real}}, x_{k} + B_{\mathrm{real}}, u_{k} + w_{k},
\
y_{k} = C_{\mathrm{real}}, x_{k} + v_{k}
\end{equation}
where $w_{k}\in\mathbb{R}^{n}$ and $v_{k}\in\mathbb{R}^{q}$ represent stochastic disturbances, and $A_{\mathrm{real}}, B_{\mathrm{real}}, C_{\mathrm{real}}$ are the true system matrices. Unlike the classical observer design scenario, where the system matrices are assumed to be known, in our setting, only nominal (possibly noisy) estimates
\begin{equation}
\label{eq:nominal}
A = A_{\mathrm{real}} - \delta_{A}, \
B = B_{\mathrm{real}} - \delta_{B}, \
C = C_{\mathrm{real}} - \delta_{C}
\end{equation}
are available, where the deviations $\delta_{A}, \delta_{B}, \delta_{C}$ are unknown but modest. Substituting~\eqref{eq:nominal} into~\eqref{eq:noisyTrueSystem} yields
\begin{equation}
\label{eq:noisyExpanded}
x_{k+1} = (A+\delta_{A}) x_{k} + (B+\delta_{B}) u_{k} + w_{k},
\
y_{k} = (C+\delta_{C}) x_{k} + v_{k}
\end{equation}

The general objective is to construct an observer
\begin{equation}
    \hat{x}_{k+1} = f(\hat{x}_k, u_k, y_k)
\end{equation}
that guarantees
\begin{equation}
    \lim_{k \to \infty} \| x_k - \hat{x}_k \|_{1} = 0
\end{equation}
where $\|\bullet\|_{1}$ denotes the 1-norm for a vector as well as for a matrix.

When the nominal parameters coincide with the real ones, the standard Luenberger observer,
\begin{equation}
\label{eq:luenbergerIdeal}
\hat{x}_{k+1} = A \hat{x}_k + B u_k + L ( y_k - C \hat{x}_k )
\end{equation}
achieves this objective by choosing $L$ such that $A - LC$ is Schur, with auxiliary optimal requirements \cite{Bernstein1989Steady,Douglas1991Process,Edelmayer1994An,Edelmayer1996H}.

However, when $(A,B,C)$ differ from the true matrices, as in~\eqref{eq:noisyExpanded}, the estimation error dynamic of $e_{k}:=x_{k}-\hat{x}_{k}$ become 
\begin{equation}
\label{eq:errorsystem}
    e_{k+1}=(A-LC)e_{k}+\delta_{A}x_{k}+\delta_{B}u_{k}+w_{k}-Lv_{k}+L\delta_{C}x_{k}
\end{equation}
which may fail to converge even if $A-LC$ is stable. As such, the key challenge addressed in this work is to design a more robust Luenberger-style observer for the system in Eq. (\ref{eq:noisyExpanded}).

We assume throughout that the system in Eq.~(\ref{eq:noisyExpanded}) is observable and Schur stable, and that the parameter deviations $\delta_A,\delta_B,\delta_C$ are modest. These assumptions are standard for the observer design problem. Also, without loss of generality, assume that $p\leq n$.

\section{Learning-Enhanced Luenberger Observer}
\label{sec:method}

This section presents the proposed \textit{Learning-Enhanced Observer} (LEO) framework, which improves state estimation accuracy when only noisy system parameters are available. The central idea is to treat the nominal matrices $(A,B,C)$ as optimization variables, quantify the mismatch between the observer output and the true system output, and iteratively refine the system parameters using gradient-based learning. After convergence, a Luenberger observer is reconstructed using the optimized model.

\subsection{From a Noisy LTI System to a Slowly-Varying LTV Model}

Consider the noisy LTI system in~\eqref{eq:noisyExpanded}. For each $k$, there exist matrices $A_{w,k}\in\mathbb{R}^{n\times n}$
$B_{w,k}\in\mathbb{R}^{n\times q}$ and $C_{v,k}\in\mathbb{R}^{q\times n}$ such that
\begin{equation}
    A_{w,k} x_k + B_{w,k} u_k=w_{k},\ C_{v,k}x_{k}=v_{k}
\end{equation}
Therefore
\begin{subequations}
\label{eq:LTV}
\begin{align}
&x_{k+1} = A_k x_k + B_k u_k, \ y_k = C_k x_k, \\
&A_k = A + \delta_A+ A_{w,k},\\
&B_k = B + \delta_B + B_{w,k},\\
&C_k = C + \delta_C + C_{v,k}
\end{align}
\end{subequations}
When the noise is modest, the matrices $(A_{k},B_{k},C_{k})$ deviate modestly from $(A,B,C)$.

\subsection{Local LTI Approximation of an LTV System}

Designing an observer for a genuinely LTV system is significantly more challenging than for an LTI system, particularly when the parameters are uncertain. To mitigate this difficulty, we employ an approximation argument: an LTV system can be locally matched exactly by an appropriately chosen LTI system.

\begin{Proposition}
\label{Prop1}
Let the LTV system in Eq. ~(\ref{eq:LTV}a) be driven by a fixed input sequence $u_k$. Define $X_{K,N} := (x_K\ x_{K+1}\ \cdots\ x_{K+N-1})$. If $X_{K,N}$ has full column rank, then there exists an LTI system:
    \begin{equation}
    \label{SimulatingLTI}
        \bar{x}_{k+1}=\bar{A}\bar{x}_{k}+\bar{B}\bar{u}_{k},\ \bar{y}_{k}=\bar{C}\bar{x}_{k},\ \bar{x}_{K}=x_{K}
    \end{equation}
    such that
    \begin{equation}
        \bar{x}_{j}=x_{j},\ \bar{y}_{j}=y_{j},\ j=K,K+1,...,K+N-1
    \end{equation}
\end{Proposition}

\textbf{Sketch Proof:} For the system in Eq. (\ref{eq:LTV}a), let $X'_{K,N}:=(x_{K+1}\ x_{K+2}\ ...\ x_{K+N})$, $U_{K,N}:=(u_{K}\ u_{K+1}\ ...\ u_{K+N-1})$, $Y_{K,N}=(y_{K}\ y_{K+1}\ ... \ y_{K+N-1})$. Since $X_{K,N}$ has full column rank, 
$T:=\begin{pmatrix}
X_{K,N}\\
U_{K,N}
\end{pmatrix}$
has full column rank. Hence, a general construction of $\bar{A}$, $\bar{B}$, $\bar{C}$ follows from
\begin{equation}
    (\bar{A}\ \bar{B}):=X'_{K,N}T^{\dagger},\ \bar{C}:=Y_{K,N}X_{K,N}
\end{equation}
where $T^{\dagger}$ denotes the Moore-Penrose inverse of $T$. $\blacksquare$

In practice, if the LTV system comes from a noisy system, in particular from the LTI system in Eq. (\ref{NoiseLTIFinal}) with non-zero random noises, $X_{K,N}$ can almost always have full column rank when $N\leq n$, since non-degeneration is a generic property of square matrices. However, the requirement of the exact matching of $\bar{x}_{K}=x_{K}$ is generally unavailable. The following result allows us to circumvent this limitation.
\begin{Proposition}
\label{Prop2}
    For the LTI system described in Eq. (\ref{SimulatingLTI}) with fixed $u_{k}$, if $\bar{A}$ has full rank, then $\forall K\in\mathbb{N}_{+},\ x_{K}\in \mathbb{R}^{n}$, there exists initial condition $\bar{x}_{0}$ of the system, such that $\bar{x}_{K}=x_{K}$.
\end{Proposition}
\textbf{Sketch Proof:} 
\begin{equation}
 \bar{x}_{K}=\bar{A}^{K}\bar{x}_{0}+\sum_{i=0}^{K-1}\bar{A}^{K-1-i}Bu_{i}   
\end{equation}
If $\bar{A}$ is invertible, then 
\begin{equation}
\bar{x}_{0}=\bar{A}^{-K}(\bar{x}_{K}-\sum_{i=0}^{K-1}\bar{A}^{K-1-i}Bu_{i})   
\end{equation}
satisfies the requirement. $\blacksquare$

Proposition \ref{Prop2} requires that $\bar{A}$ has full rank, which is non-trivial. However, since invertibility is a generic property of matrices, any non-invertible $\bar{A}$ can be arbitrarily well-approximated by a nearby invertible matrix, as stated below.
\begin{Lemma}
\label{Lemma1}
    $\forall\ \bar{A}\in Mat_{n\times n}(\mathbb{R}),\ \delta>0,\ \exists\ \tilde{A}\in Mat_{n\times n}(\mathbb{R})$ which is inevitable, such that $\|\bar{A}-\tilde{A}\|_{1}<\delta$.
\end{Lemma}

\textbf{Sketch Proof:} The measure of the set formed by degenerated matrices is 0, while the measure of the ball neighborhood $\{\tilde{A}\ |\ \|\bar{A}-\tilde{A}\|_{1}<\delta\}$ is larger than 0. $\blacksquare$

Consequently, one obtains the following approximation guarantee.

\begin{Corollary}
    $\forall\ \epsilon>0$, the LTV system in Proposition \ref{Prop1} admits an approximating LTI system 
    \begin{equation}
        \label{SimulatingLTIFinal}
        \bar{x}_{k+1}=\bar{A}\bar{x}_{k}+\bar{B}\bar{u}_{k},\ \bar{y}_{k}=\bar{C}\bar{x}_{k}
    \end{equation}
 such that $\|x_{k}-\bar{x}_{k}\|_{1}<\epsilon$ for $k\in\{K,K+1,...,K+N-1\}$.
\end{Corollary}

\textbf{Sketch Proof:} For an LTI system described in Eq. (\ref{SimulatingLTIFinal}) and a fixed $k$, $\bar{x}_{k}$ is continuous as a map of $\bar{A}$, and thus $\exists\ \delta_{i}$ such that $\|\bar{x}_{i}-\tilde{x}_{i}\|_{1}<\epsilon$ whenever $\|\bar{A}-\bar{A}_{i}\|_{1}<\delta_{i}, \ i=K,K+1,...,K+N-1$, where $\bar{x},\ \tilde{x}$ are states of system $(\bar{A},\bar{B})$ and $(\bar{A}_{i},\bar{B})$, respectively. Let $\delta=min_{i}\{\delta_{i}\}$ which exists since $N$ is finite. Select an invertible matrix $\tilde{A}$ with $\|\bar{A}-\tilde{A}\|_{1}<\delta$, which exists by Lemma \ref{Lemma1}, then the system formed by $(\tilde{A},\bar{B})$ satisfies the requirements. $\blacksquare$

Still, the initial condition and the system parameters in Eq. (\ref{SimulatingLTIFinal}) are unknown. However, the problem is reduced to the state estimation for the LTI system in Eq. (\ref{SimulatingLTIFinal}), which motivates the usage of an LTI observer to estimate the states of the noisy system in Eq. (\ref{eq:noisyExpanded}). Hence, a Luenberger observer can be used.

\subsection{Learning-Based Refinement of Luenberger Observers}

Returning to the original problem~(\ref{eq:noisyExpanded}), we use a Luenberger-type structure but treat $(A,B,C,x_{0})$ as learnable variables. A standard observer is given in Eq. (\ref{eq:luenbergerIdeal}). The LEO algorithm introduces a loss function that quantifies the steady-state output discrepancy between the observer and the true system, which is used to update the observer parameters:
\begin{equation}
\label{Loss}
\begin{aligned}
    &\mathcal{L}(A, B, C,\hat{x}_{0}) 
    \\
    =& \frac{1}{K} \sum_{k=k_0}^{k_0+K} \| y_k - C \hat{x}_k \| + \lambda_{A} \mathcal{R}(A) + \lambda_{B} \mathcal{R}(B) + \lambda_{C} \mathcal{R}(C)
\end{aligned}
\end{equation}
where $k_{0}$ is a chosen threshold, $K$ is the averaging window, $\| X \|$ is the average of absolute value of the elements in $X$ for a vector or a matrix, $\lambda_{A},\lambda_{B},\lambda_{C}$ are regularization coefficients, and 
\begin{equation}
    \mathcal{R}(A):=\| A-A_{init} \|,\ \mathcal{R}(B):=\| B-B_{init} \|,\ \mathcal{R}(C):=\| C-C_{init} \|
\end{equation} 
are the regularization terms (see the next subsection). The parameters $(A, B, C,\hat{x}_{0})$ are then updated using a gradient-based optimizer, initially from the nominal estimation of $(A_{real}, B_{real}, C_{real})$ and a initial state estimation $\hat{x}_{0,init}$.
\begin{subequations}
\begin{align}
&(A, B, C,\hat{x}_{0}) \gets (A, B, C,\hat{x}_{0}) - \eta \nabla \mathcal{L}(A, B, C,\hat{x}_{0})
\\
&(A, B, C,\hat{x}_{0})_{init}=(A_{real}-\delta_A, B_{real}-\delta_B, C_{real}-\delta_C,\hat{x}_{0,init})
\end{align}
\end{subequations}
where $\eta$ is the learning rate. After convergence, the optimized parameters $(A_{\mathrm{opt}}, B_{\mathrm{opt}}, C_{\mathrm{opt}},x_{0,\mathrm{opt}})$ are used to reconstruct the final Luenberger observer:
\begin{equation}
\hat{x}_{k+1}^{\mathrm{opt}} = A_{\mathrm{opt}} \hat{x}_k^{\mathrm{opt}} + B_{\mathrm{opt}} u_k + L_{\mathrm{opt}} (y_k - C_{\mathrm{opt}} \hat{x}_k^{\mathrm{opt}}).
\end{equation}

\subsection{Practical Considerations}

The optimization-based refinement of the observer parameters introduces several practical issues that must be handled. The LEO framework incorporates safeguards addressing invariance under similarity transformations, numerical stability of the observer design, and possible loss of observability during learning.

\subsubsection{Invariance and Regularization Against Similarity Drift}

Matching the observer output $\hat{y}_k = C\hat{x}_k$ to the true output $y_k$ does not uniquely determine the internal state trajectory or system realization, since LTI systems are invariant under similarity transformations. However,
\begin{Proposition}
\label{Prop3}
    Let two LTI systems
    \begin{subequations}
    \begin{align}
        &\Sigma_{1}: x^{1}_{k}=A_{1}x^{1}_{k}+B_{1}u_{k},\  y^{1}_{k}=C_{1}x^{1}_{k}
        \\
        &\Sigma_{2}: x^{2}_{k}=A_{2}x^{2}_{k}+B_{2}u_{k},\  y^{2}_{k}=C_{2}x^{2}_{k}
    \end{align} 
    \end{subequations} share the same input and output sequences $\{u_k,y_k\}_{k=0}^{N-1}$. Define for $i=1,2$:
    \begin{equation}
    \begin{aligned}
        &O^{i}_{N}:= \begin{pmatrix}
 C_{i}A_{i}^{0}\\
 C_{i}A_{i}^{1}\\
 ...\\
C_{i}A_{i}^{N-1}
\end{pmatrix},\
U_{N-1}:= \begin{pmatrix}
 u_{0}\\
 u_{1}\\
 ...\\
u_{N-2}
\end{pmatrix},
\\
&\Gamma^{i}_{N}:= \begin{pmatrix}
0 & 0 &...& 0\\
 C_{i}A_{i}^{0}B_{i} & 0 &...& 0\\
 ...&...&...&...\\
C_{i}A_{i}^{N-2}B_{i} &C_{i}A_{i}^{N-3}B_{i} &...& C_{i}A_{i}^{0}B_{i}
\end{pmatrix}
    \end{aligned}
    \end{equation}
If $O^{1}_{N}$ has full column rank, then
\begin{equation}
\begin{aligned}
    &\quad \|x^{1}_{0}-x^{2}_{0}\|_{1}
    \\
    &\leq \|(O^{1}_{N})^{\dagger}\|_{1}(\|\Gamma^{1}_{N}-\Gamma^{2}_{N}\|_{1}\|U_{N-1}\|_{1}+\|O^{1}_{N}-O^{2}_{N}\|_{1}\|x^{2}_{0}\|_{1})
\end{aligned}
\end{equation}
where $(O^{1}_{N})^{\dagger}$ represents the Moore–Penrose pseudoinverse of $O^{1}_{N}$.
\end{Proposition}

\textbf{Proof:} For $i=1,2$, the first $N$ outputs can be written as:
\begin{equation}
    Y^{i}_{N}=O^{i}_{N}x^{i}_{0}+\Gamma^{i}_{N}U_{N-1}
\end{equation}
where
\begin{equation}
    Y^{i}_{N}:=\begin{pmatrix}
 y^{i}_{0}\\
 y^{i}_{1}\\
 ...\\
y^{i}_{N-1}
\end{pmatrix}
\end{equation}
Hence, the equality of outputs implies that
\begin{equation}
    O^{1}_{N}x^{1}_{0}+\Gamma^{1}_{N}U_{N-1}=O^{2}_{N}x^{2}_{0}+\Gamma^{2}_{N}U_{N-1}
\end{equation}
and thus, using $O^{1}_{N}$ has full column rank,
\begin{equation}
\begin{aligned}
    &\qquad \|x^{1}_{0}-x^{2}_{0}\|_{1}
    \\
    &= \|(O^{1}_{N})^{\dagger}[(\Gamma^{2}_{N}-\Gamma^{1}_{N})U_{N-1}-(O^{1}_{N}-O^{2}_{N})x^{2}_{0}]\|_{1}
    \\
    &\leq \|(O^{1}_{N})^{\dagger}\|_{1}(\|\Gamma^{1}_{N}-\Gamma^{2}_{N}\|_{1}\|U_{N-1}\|_{1}+\|O^{1}_{N}-O^{2}_{N}\|_{1}\|x^{2}_{0}\|_{1})
\end{aligned}
\end{equation}
$\blacksquare$

Proposition \ref{Prop3} demonstrates that if two LTI systems have close system parameters, bounded inputs and states, and $\Gamma_{1}$ is observable with a non-very-low singular value, which implies that $\|(O^{1}_{N})^{\dagger}\|_{1}$ is small, then the difference of their states can be restricted. 

This motivates the regularization terms $\mathcal{R}(A),\mathcal{R}(B),\mathcal{R}(C)$ in the loss function Eq. (~\ref{Loss}), which constrain the learned parameters to remain close to their nominal values. Under the assumption that parameter deviations are modest, this prevents the optimized parameters $(A_{\mathrm{opt}},B_{\mathrm{opt}},C_{\mathrm{opt}})$ drifting far from the real ones $(A_{real},B_{real},C_{real})$, and thereby restricting the drift of the estimated states when the real system is well-conditioned.

\subsubsection{Conditions Number of the Observability Matrices.}  

The Luenberger gain $L$ is computed by pole placement on the error system Eq. (\ref{eq:errorsystem}). If the observability matrix associated with $(A,C)$ has a high condition number, the pole placement algorithm can return a large gain $L$, resulting in numerical instability and noise amplification.

To limit this effect, we apply a similarity transformation on the estimation system, ensuring that the observability matrix is well-conditioned. Optimization is then performed in the normalized coordinates, and the inverse transformation is applied after learning to recover the updated matrices in the original basis.

\subsubsection{Observability During Learning}

Gradient-based updates occasionally yield temporary loss of observability, making the computation of $L$ infeasible. Since unobservable parameter sets have measure zero, this event is rare. When it does occur, the observer temporarily reuses the previous gain $L$, ensuring optimization can continue. In practice, this safeguard is sufficient to maintain stable learning dynamics and rarely affects the accuracy.

\subsection{Algorithm Summary}

In summary, the LEO algorithm is as follows:
The complete LEO procedure is summarized below.

\begin{breakablealgorithm}
\caption{LEO algorithm with Luenberger observer design}
\begin{algorithmic}[1]
\STATE \textbf{Input:} nominal parameters $(A_{\mathrm{init}},B_{\mathrm{init}},C_{\mathrm{init}})$, initial estimate $\hat{x}_{0,\mathrm{init}}$, input-output data ${u_k,y_k}$
\STATE \textbf{Learnable parameters:} $(\hat{A},\hat{B},\hat{C},\hat{x}_{0})$
\STATE \textbf{Initialize:} $(\hat{A},\hat{B},\hat{C},\hat{x}_{0})\gets(A_{\mathrm{init}},B_{\mathrm{init}},C_{\mathrm{init}},\hat{x}_{0,\mathrm{init}})$
\REPEAT
\STATE Apply equivalent transformation to enhance the condition number if necessary
\STATE Design Luenberger observer with current parameters $(\hat{A},\hat{B},\hat{C},\hat{x}_{0})$: If $(\hat{A},\hat{C})$ is observable, update $L$ using pole placement, else retain previous $L$
\STATE Compute loss $\mathcal{L}(\hat{A},\hat{B},\hat{C},\hat{x}0)$ as in~\eqref{Loss}
\STATE Update parameters using gradient descent:
\UNTIL{convergence}
\STATE \textbf{Output:} optimized parameters $(A_{\mathrm{opt}},B_{\mathrm{opt}},C_{\mathrm{opt}},\hat{x}_{0,\mathrm{opt}})$
\STATE Design the final enhanced-observers via the optimized parameters $(A_{\mathrm{opt}},B_{\mathrm{opt}},C_{\mathrm{opt}},\hat{x}_{0,\mathrm{opt}})$
\end{algorithmic}
\end{breakablealgorithm}

By combining classical Luenberger observer design with learning-based parameter optimization, the LEO framework achieves robust state estimation with reduced error, even in the presence of parameter uncertainty and noise.

\section{Numerical Experiments}

This section evaluates the proposed LEO through extensive numerical simulations. Both open-loop estimation and closed-loop Luenberger observers are examined under randomly generated systems, disturbances, and initialization uncertainties.

\subsection{Experimental Setup}

Learning is performed using the Adam optimizer with a weight decay of $10^{-5}$. The learning rate is initialized at $10^{-4}$ and reduced by a factor of $10$ every $200$ epochs, over a total of 250 epochs. The loss function employs a steady-state window starting at $k_0=201$ with window size $K=50$. Regularization coefficients follow
\begin{equation}
    \lambda_A=10^{-3}\frac{n^2}{n^2+np+nq},\ 
\lambda_B=10^{-3}\frac{np}{n^2+np+nq},\ 
\lambda_C=10^{-3}\frac{nq}{n^2+np+nq}
\end{equation}

The normalized error is used to measure the deviation of the estimated states from the real ones, defined as:
\begin{equation}
    e=\|\frac{\hat{x}_{k}-x_{k}}{x_{k}}\|
\end{equation}
where the vector division $\frac{a}{b}$ here is understood as component-wise division.

\subsection{Statistical Results}

Table~\ref{Table1} reports the Monte Carlo results for systems of dimension $n=2$ to $n=4$, with input/output dimensions satisfying $p \geq \lfloor n/2 \rfloor$, $q \leq p$, and $q < n$. For each system dimension $(n,p,q)$, we conduct $100$ randomized trials. In every trial, a random LTI system is generated, and independent Gaussian samples are used to construct the inputs, disturbances, and parameter perturbations. 
 
We summarize three statistics:
\begin{itemize}
    \item \textbf{ERR}: Average percentage reduction in steady-state normalized error, with the top and bottom 10\% trials removed.
    \item \textbf{SR}: Success rate, i.e., the proportion of trials in which the LEO outperforms the nominal one.
    \item \textbf{p-value}: Wilcoxon signed-rank test comparing the two observers (values below $0.05$ indicate statistically significant improvement).
\end{itemize}

Across all configurations, the LEO consistently outperforms the nominal observer in both open-loop and closed-loop settings. On average, the enhanced observer achieves more than 15\% reduction in steady-state error, with success rates exceeding 70\% in nearly all cases. Furthermore, all p-values remain below $1.1\times 10^{-3}$, confirming the statistical significance of the improvements. These results indicate that gradient-based parameter refinement effectively mitigates the impact of initial model mismatch.

\begin{table}[htbp]
\renewcommand{\arraystretch}{1.5} 
\centering
\caption{Monte Carlo performance comparison of the nominal observers and the LEO. Process noise $w_k$ and measurement noise $v_k$ follow are sampled independently and identically distributed (i.i.d.) from $\mathcal{N}(0,0.01I_n)$ and $\mathcal{N}(0,0.01I_q)$, respectively. Parameter perturbations $\delta_A$, $\delta_B$, and $\delta_C$ are sampled i.i.d. from $\mathcal{N}(0,0.05^2)$, per element. The true initial state $x_{0,real}$ is drawn from $\mathcal{N}(0,1)$ per component, while the initial estimated state uses $\mathcal{N}(x_{0,real},10^2)$. Inputs are sampled from $\mathcal{N}(0,1)$.}
\label{Table1}
\begin{tabular}{|p{7mm}<{\centering}|p{35.75mm}<{\centering}|p{35.75mm}<{\centering}|}
\hline
 & Open-loop & Closed-loop 
\end{tabular}
\begin{tabular}{|p{7mm}<{\centering}|p{8mm}<{\centering}|p{5mm}<{\centering}|p{14mm}<{\centering}|p{8mm}<{\centering}|p{5mm}<{\centering}|p{14mm}<{\centering}|}
\hline
(n,p,q) & ERR & SR & p-value & ERR & SR & p-value\\
\hline
(2,1,1) & 16.19\% & 84\% & $2.3\times 10^{-9}$ & 15.14\% & 83\% & $1.9\times 10^{-9}$\\ 
\hline
(2,2,1) & 19.83\% & 81\% & $3.2\times 10^{-10}$ & 19.67\% & 82\% & $1.6\times 10^{-10}$\\ 
\hline
(3,1,1) & 25.24\% & 82\% & $2.6\times 10^{-9}$ & 21.23\% & 77\% & $2.1\times 10^{-7}$\\ 
\hline
(3,2,1) & 27.58\% & 79\% & $2.0\times 10^{-8}$ & 25.43\% & 79\% & $3.8\times 10^{-7}$\\ 
\hline
(3,2,2) & 14.87\% & 72\% & $1.1\times 10^{-5}$ & 34.19\% & 84\% & $3.1\times 10^{-12}$\\ 
\hline
(3,3,1) & 27.79\% & 84\% & $5.0\times 10^{-10}$ & 25.82\% & 83\% & $9.9\times 10^{-10}$\\ 
\hline
(3,3,2) & 15.87\% & 72\% & $7.0\times 10^{-6}$ & 39.90\% & 91\% & $8.5\times 10^{-15}$\\ 
\hline
(4,2,1) & 16.45\% & 76\% & $1.4\times 10^{-6}$ & 20.54\% & 76\% & $2.0\times 10^{-5}$\\ 
\hline
(4,2,2) & 18.08\% & 76\% & $4.2\times 10^{-6}$ & 29.95\% & 83\% & $1.9\times 10^{-8}$\\ 
\hline
(4,3,1) & 26.07\% & 80\% & $2.6\times 10^{-8}$ & 22.33\% & 77\% & $3.4\times 10^{-7}$\\ 
\hline
(4,3,2) & 23.93\% & 82\% & $1.4\times 10^{-10}$ & 32.24\% & 85\% & $4.2\times 10^{-11}$\\ 
\hline
(4,3,3) & 12.12\% & 71\% & $8.0\times 10^{-5}$ & 43.00\% & 92\% & $1.6\times 10^{-13}$\\ 
\hline
(4,4,1) & 20.60\% & 73\% & $1.1\times 10^{-4}$ & 15.92\% & 70\% & $1.1\times 10^{-3}$\\ 
\hline
(4,4,2) & 26.11\% & 82\% & $1.4\times 10^{-8}$ & 35.82\% & 87\% & $4.5\times 10^{-12}$\\ 
\hline
(4,4,3) & 17.40\% & 79\% & $1.4\times 10^{-8}$ & 45.12\% & 94\% & $1.9\times 10^{-13}$\\ 
\hline
\end{tabular}
\end{table}

\subsection{Visual Example}

\begin{figure}[htbp]
  \centering
  \begin{subfigure}[htbp]{0.48\columnwidth}
    \centering
    \includegraphics[width=\linewidth]{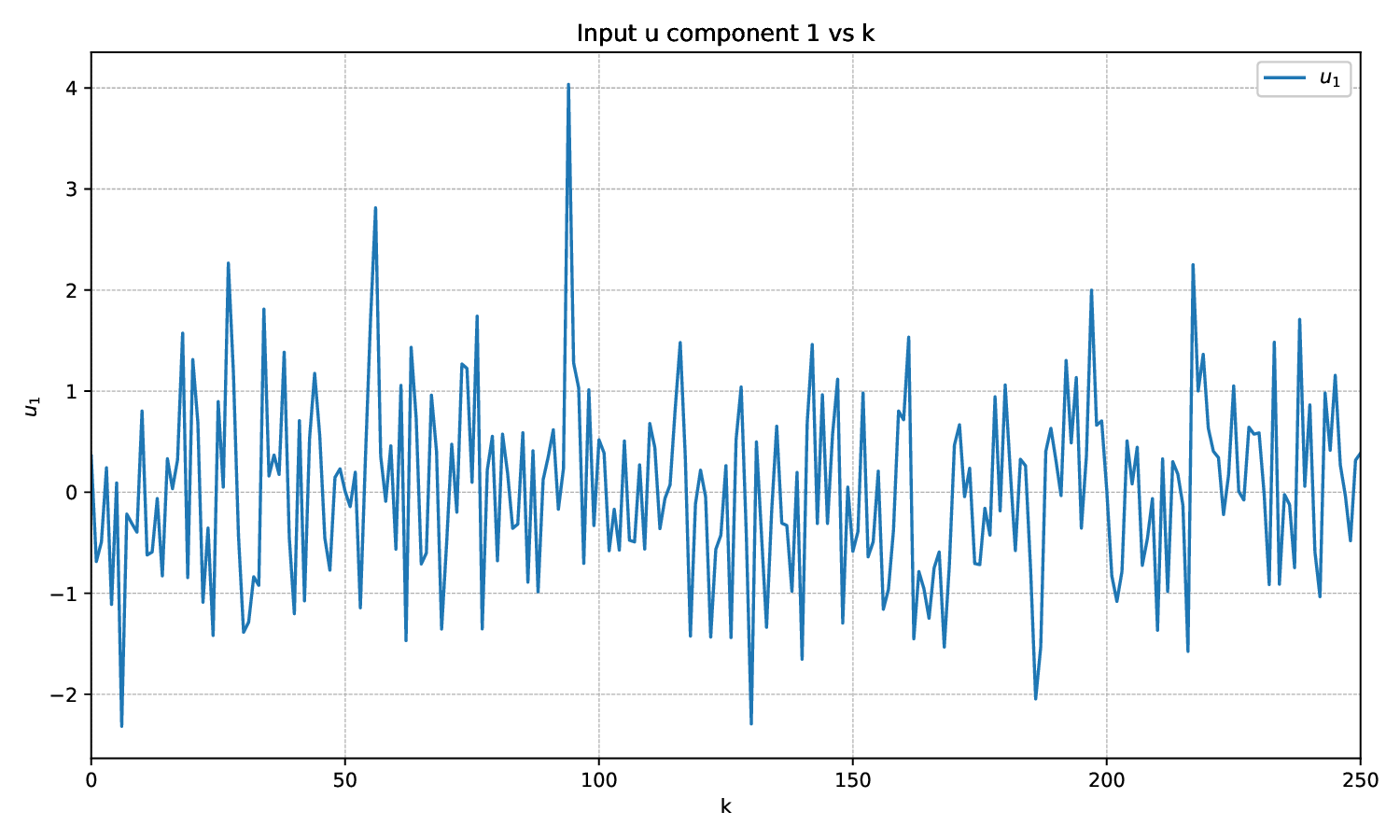}
    \caption{The input $u_{k}$ is sampled by Gaussian procedure with zero mean and standard deviation 1.}
    \label{u}
  \end{subfigure}
  \hfill
  \begin{subfigure}[htbp]{0.48\columnwidth}
    \centering
    \includegraphics[width=\linewidth]{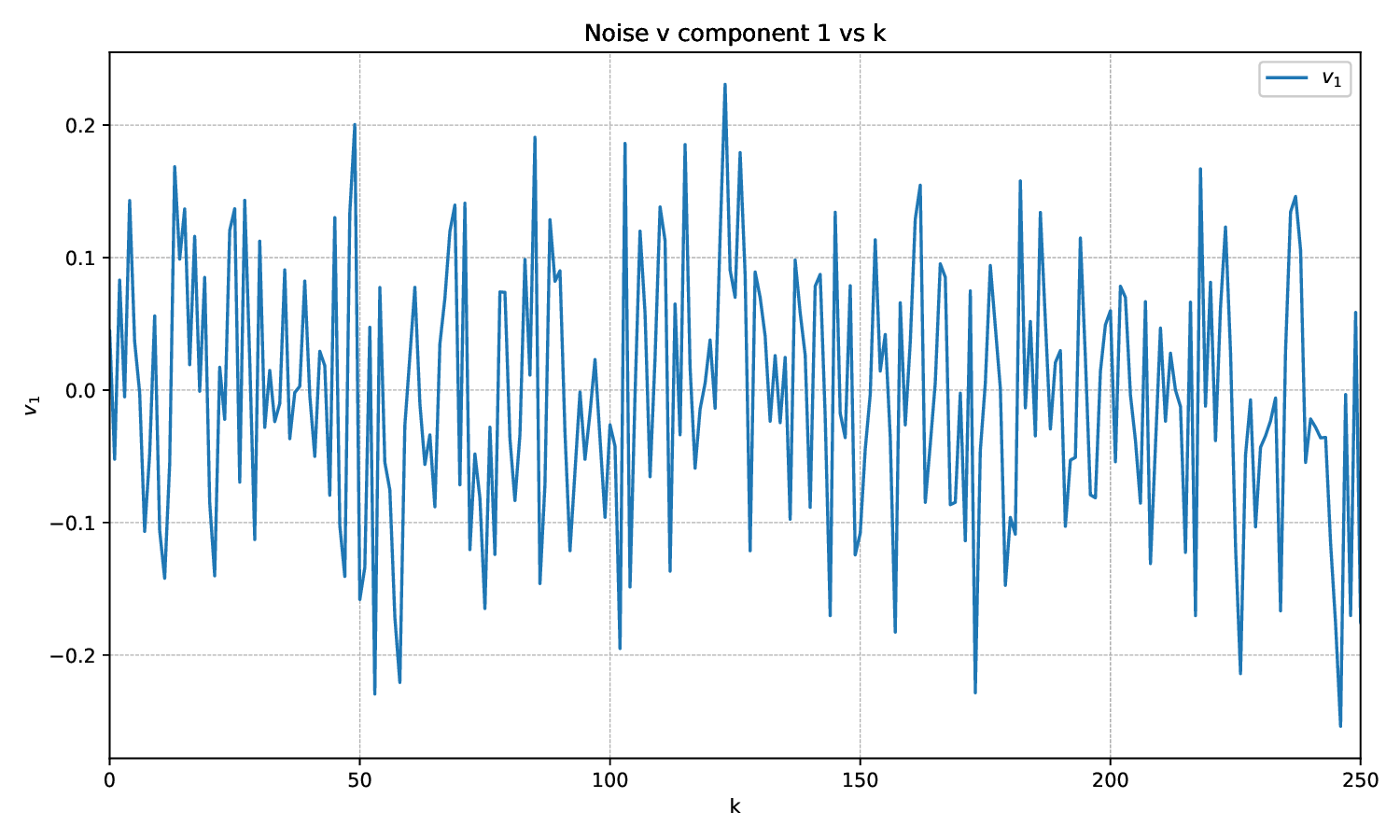}
    \caption{The noise $v_{k}$ is sampled by Gaussian procedure with zero mean and variance 0.01.}
    \label{v}
  \end{subfigure}
  \hfill
  \begin{subfigure}[htbp]{0.48\columnwidth}
    \centering
    \includegraphics[width=\linewidth]{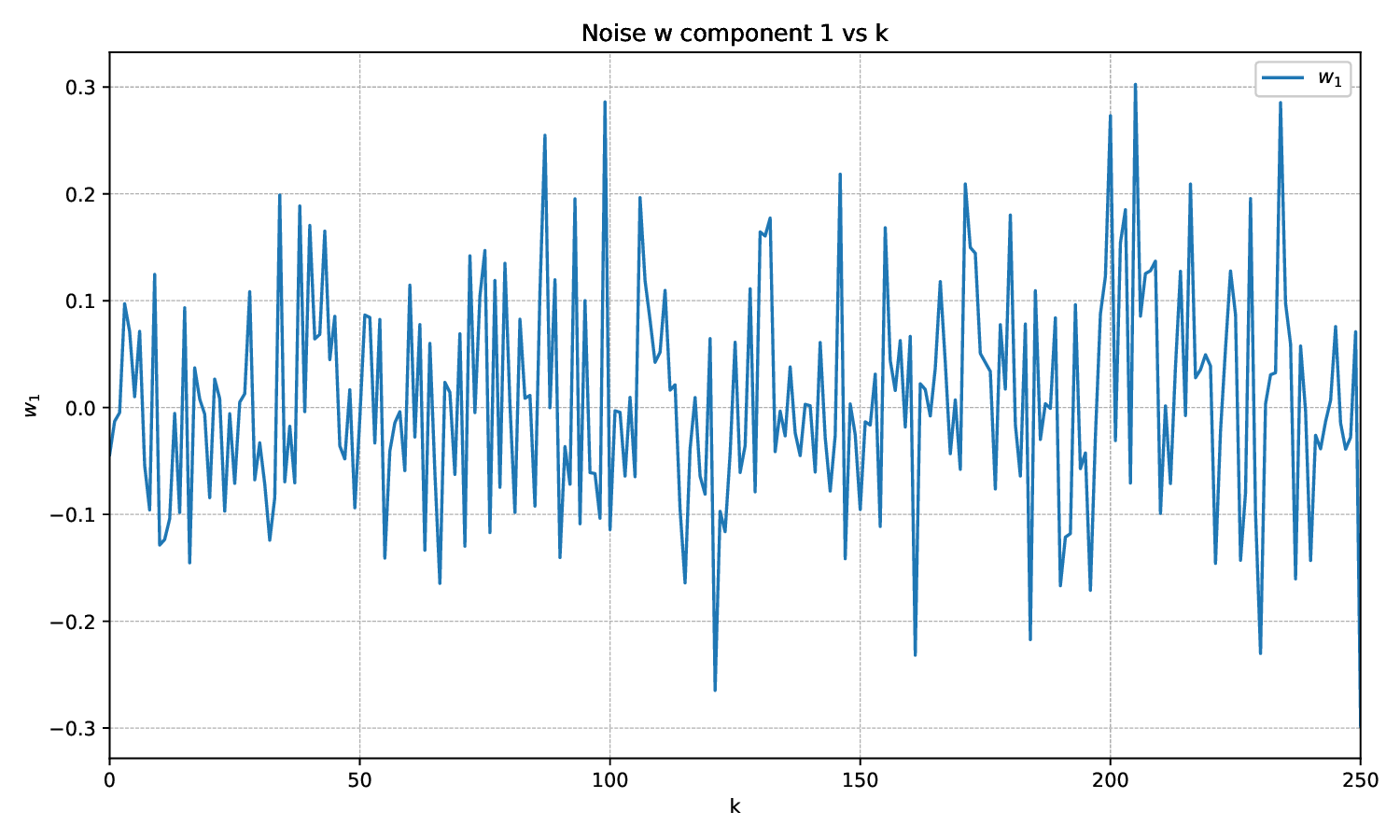}
    \caption{The noise $w$ is sampled by Gaussian procedure with zero mean and covariance matrix 0.01$I_{2\times 2}$.}
    \label{w1}
  \end{subfigure}
  \hfill
  \begin{subfigure}[htbp]{0.48\columnwidth}
    \centering
    \includegraphics[width=\linewidth]{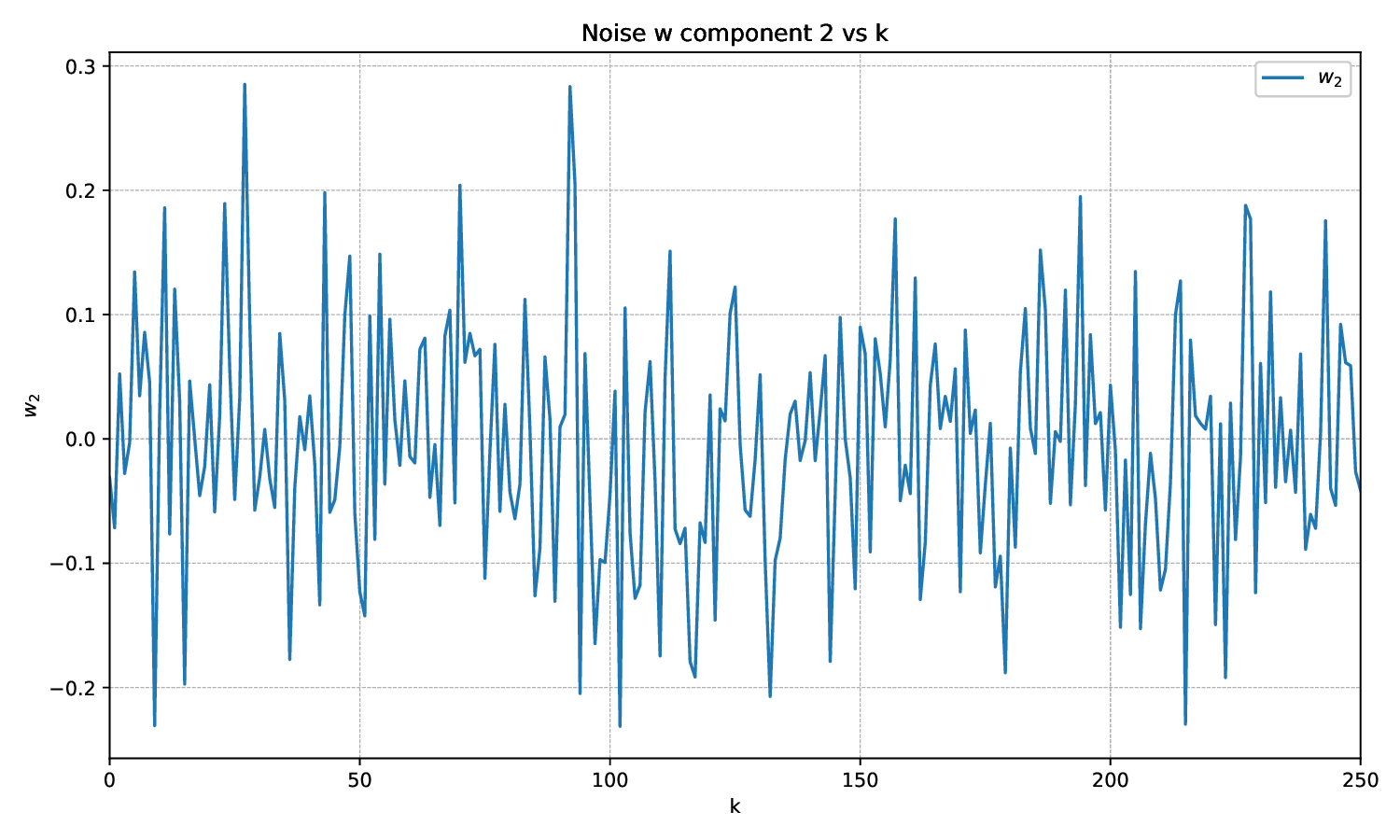}
    \caption{The noise $w$ is sampled by Gaussian procedure with zero mean and covariance matrix 0.01$I_{2\times 2}$.}
    \label{w2}
  \end{subfigure}
  \hfill
  \begin{subfigure}[htbp]{0.48\columnwidth}
    \centering
    \includegraphics[width=\linewidth]{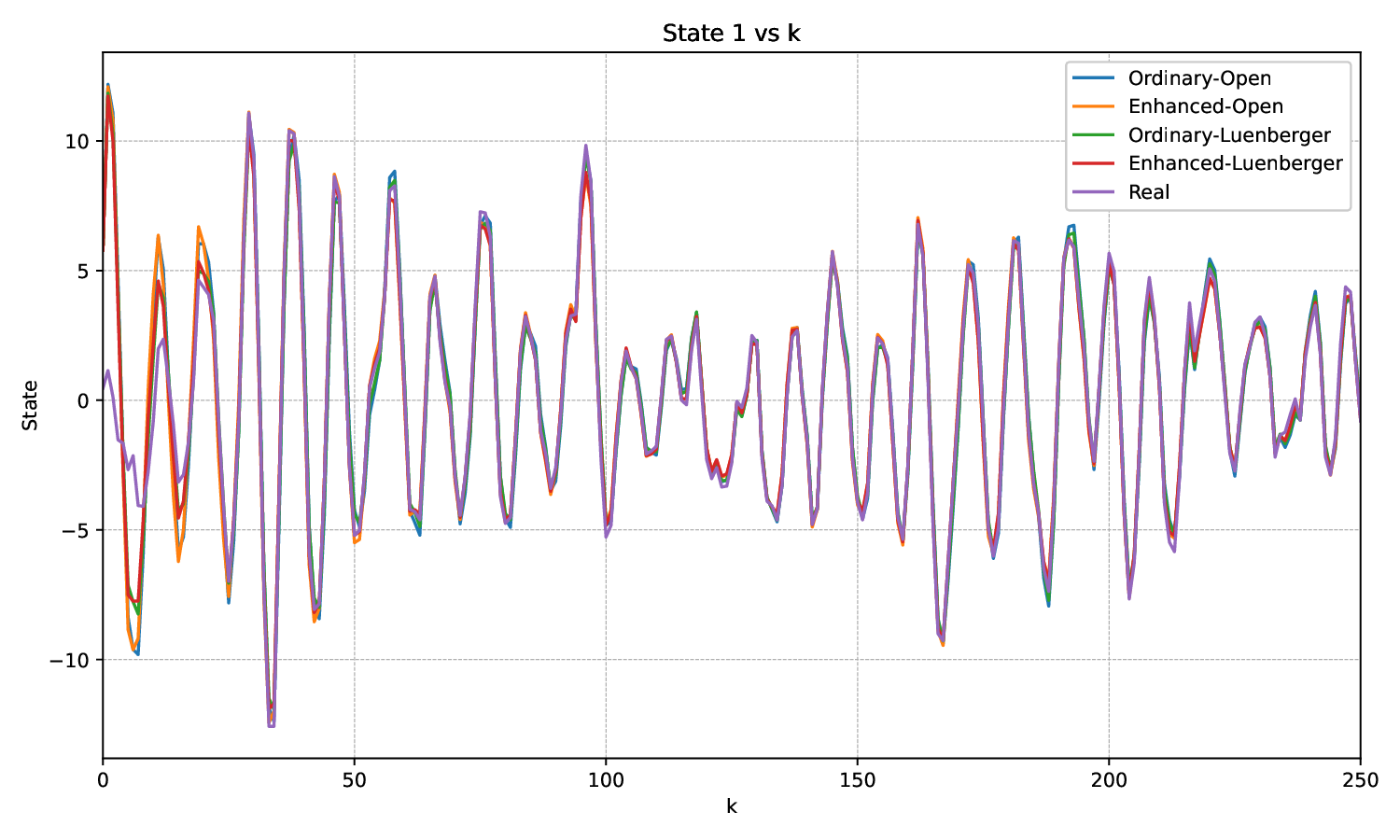}
    \caption{The first component of states vs $k$.}
    \label{x1}
  \end{subfigure}
  \hfill
  \begin{subfigure}[htbp]{0.48\columnwidth}
    \centering
    \includegraphics[width=\linewidth]{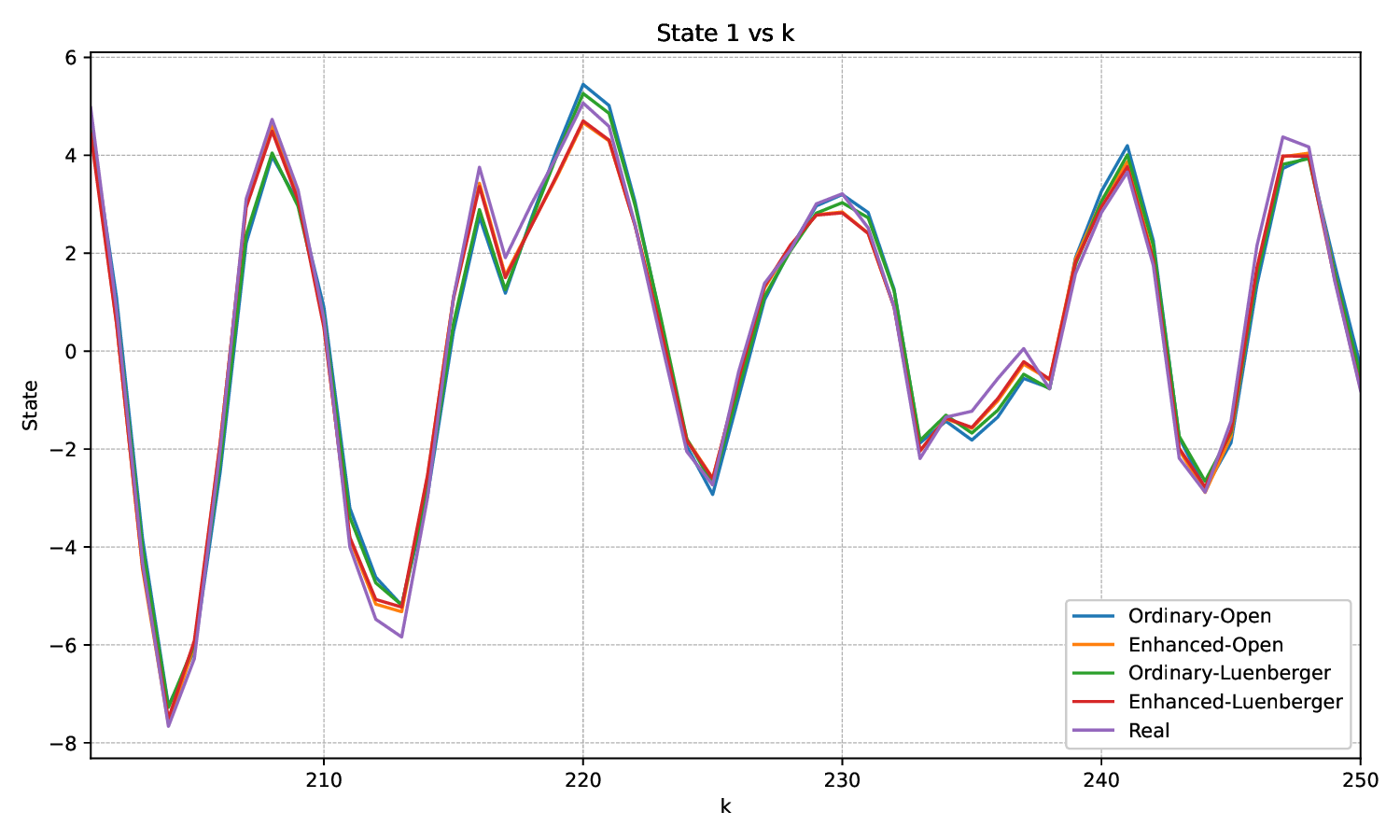}
    \caption{The first component of states vs $k$ after stable ($k$ from 201 to 251).}
    \label{stablex1}
  \end{subfigure}
  \hfill
  \begin{subfigure}[htbp]{0.48\columnwidth}
    \centering
    \includegraphics[width=\linewidth]{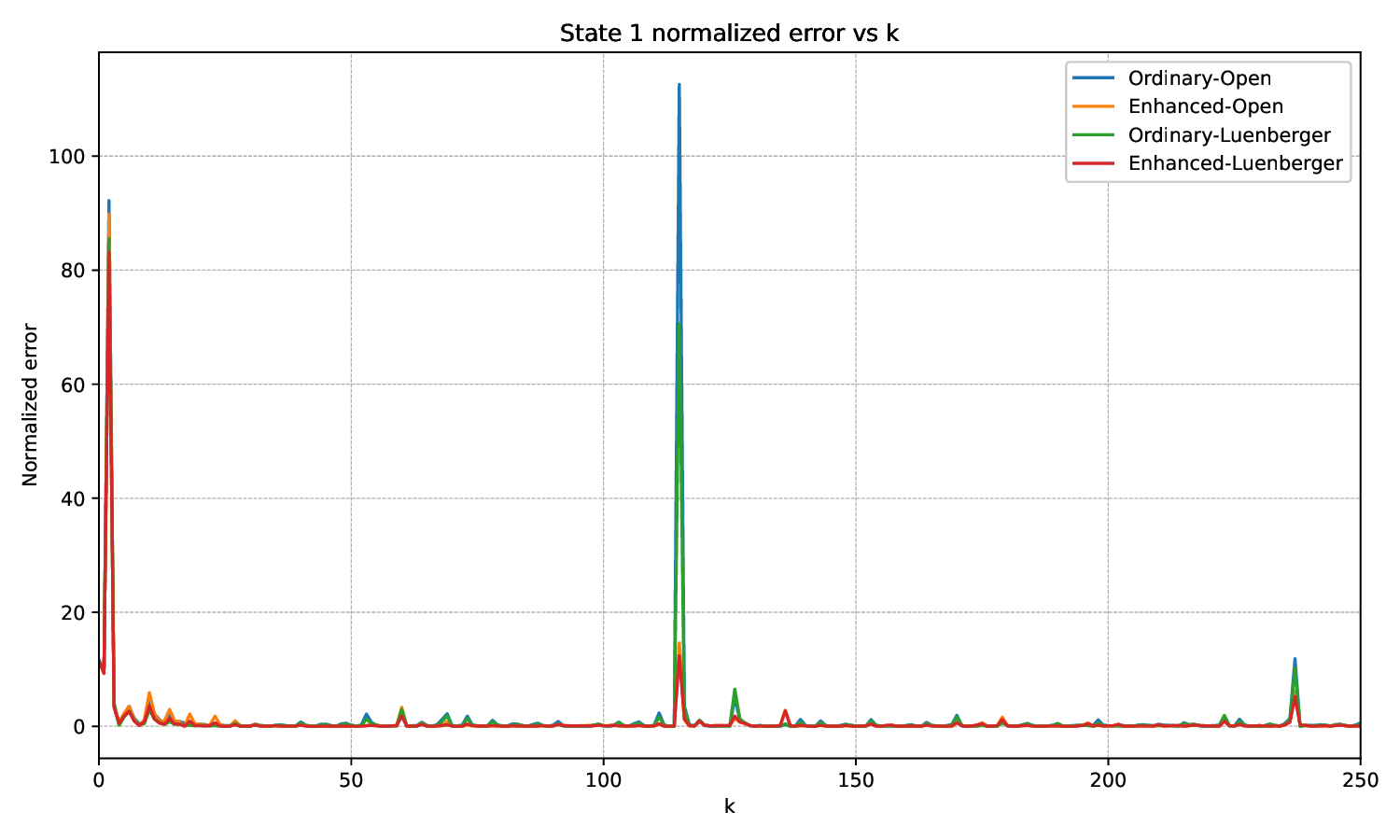}
    \caption{The first component of normalized state error vs $k$.}
    \label{e1}
  \end{subfigure}
  \hfill
  \begin{subfigure}[htbp]{0.48\columnwidth}
    \centering
    \includegraphics[width=\linewidth]{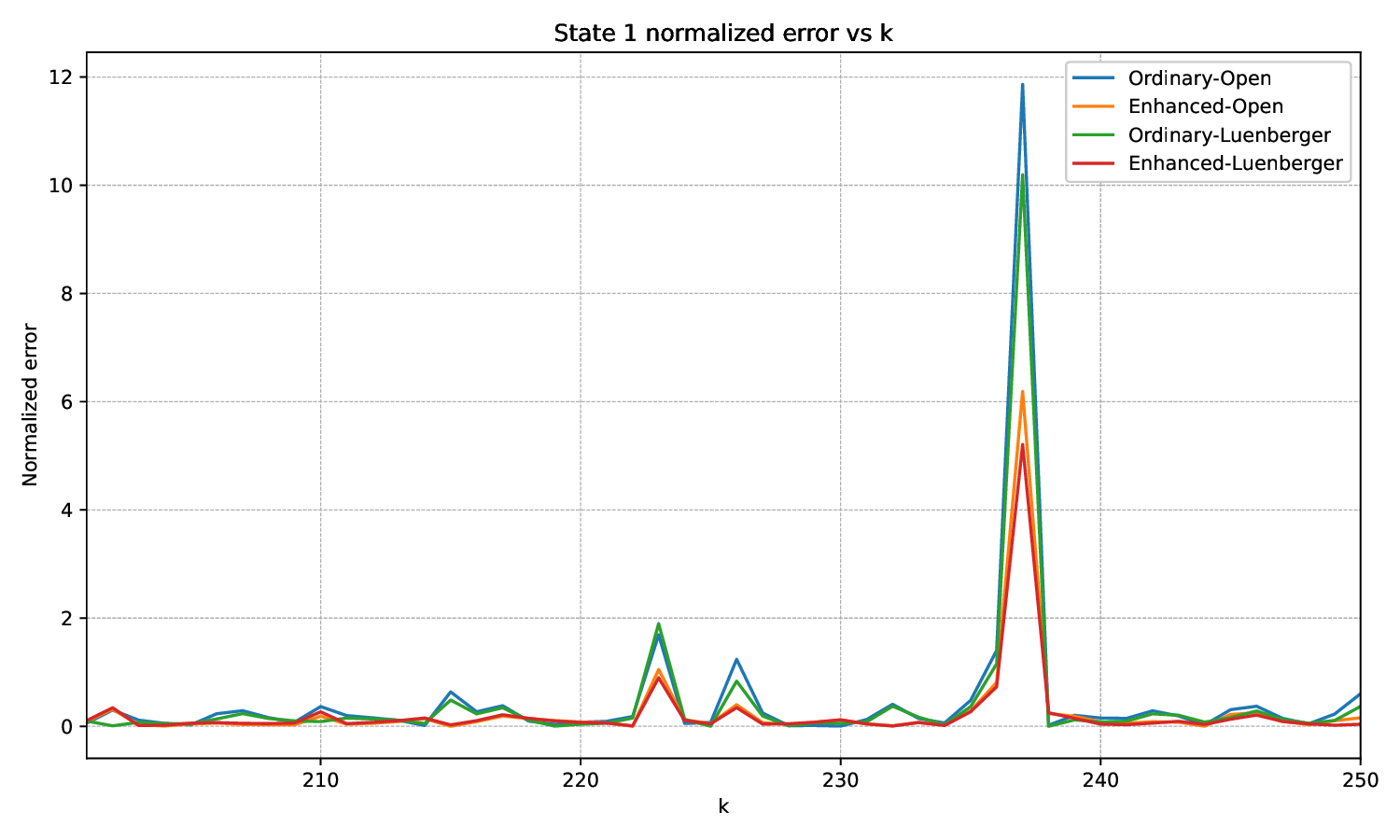}
    \caption{The first component of normalized state error vs $k$ after stable.}
    \label{stablee1}
  \end{subfigure}
  \hfill
  \begin{subfigure}[htbp]{0.48\columnwidth}
    \centering
    \includegraphics[width=\linewidth]{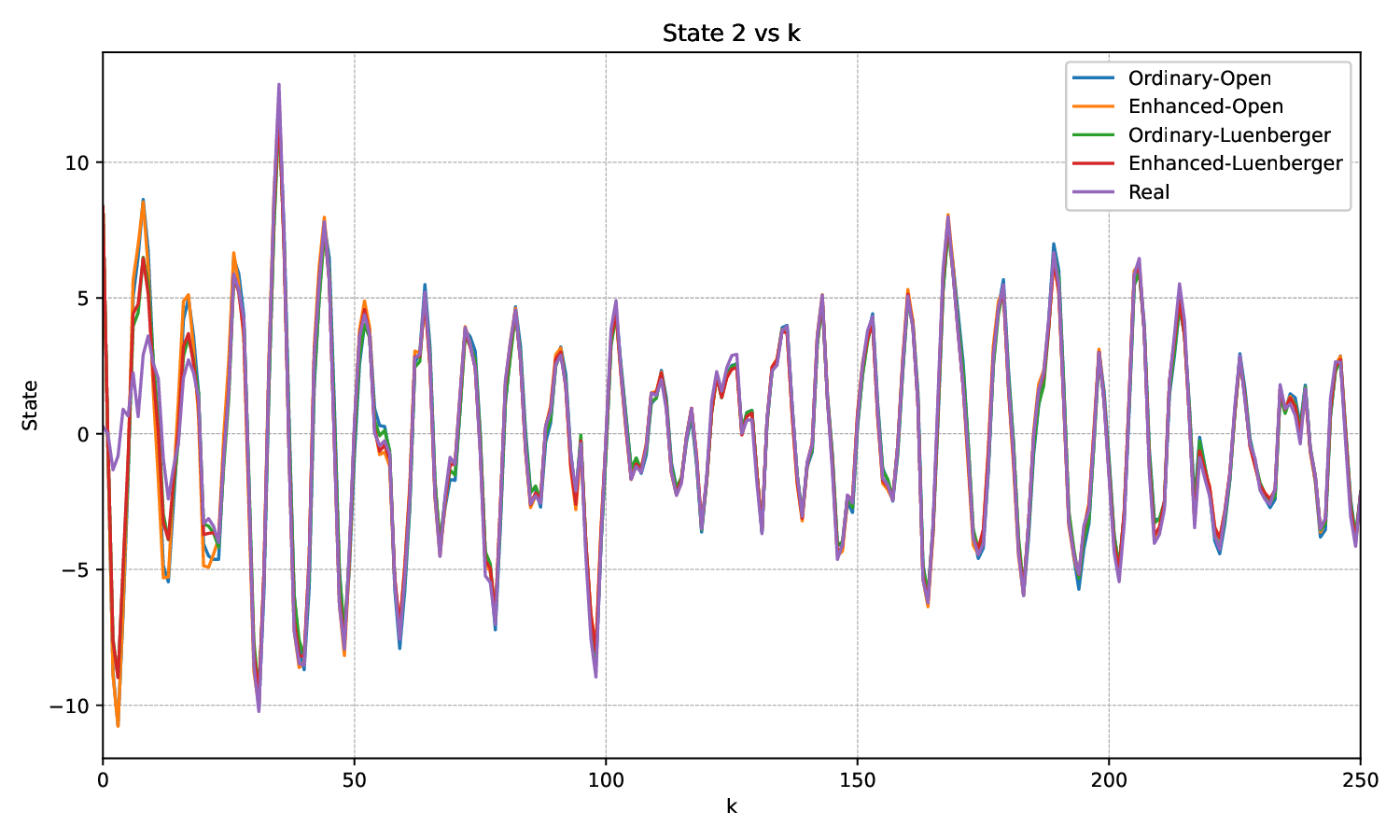}
    \caption{The second component states vs $k$.}
    \label{x2}
  \end{subfigure}
  \hfill
  \begin{subfigure}[htbp]{0.48\columnwidth}
    \centering
    \includegraphics[width=\linewidth]{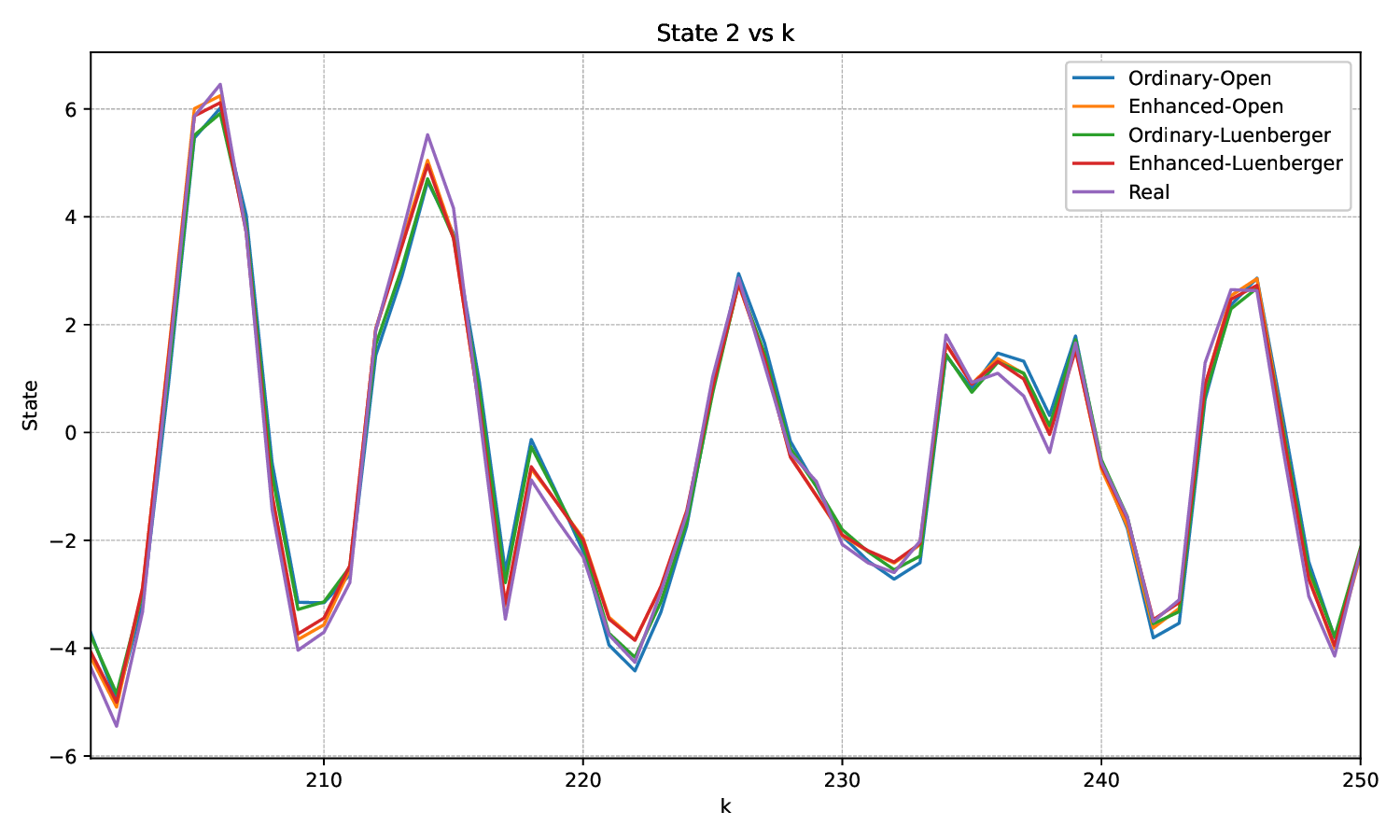}
    \caption{The second component states vs $k$ after stable.}
    \label{stablex2}
  \end{subfigure}
  \hfill
  \begin{subfigure}[htbp]{0.48\columnwidth}
    \centering
    \includegraphics[width=\linewidth]{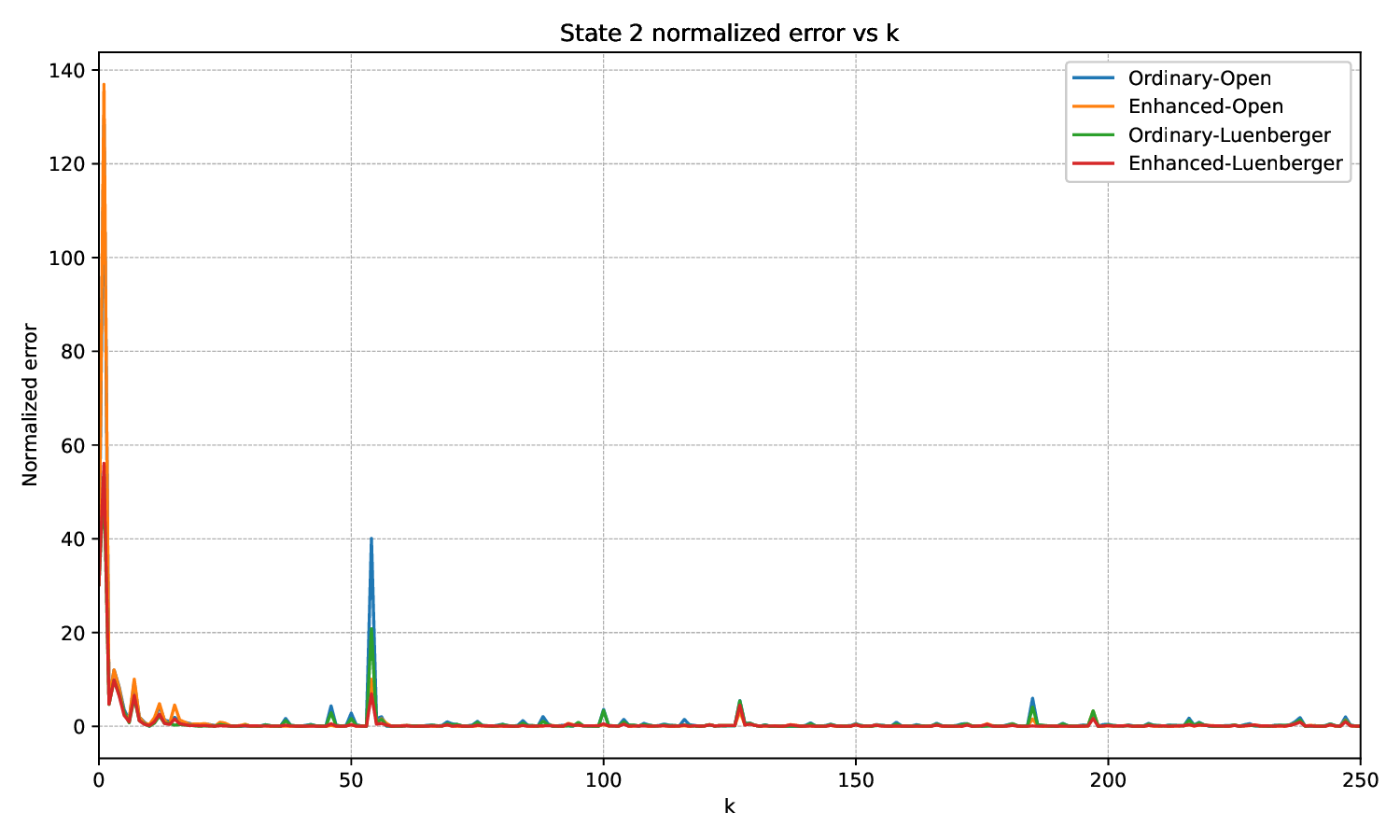}
    \caption{The second component of normalized state error vs $k$.}
    \label{e2}
  \end{subfigure}
  \hfill
  \begin{subfigure}[htbp]{0.48\columnwidth}
    \centering
    \includegraphics[width=\linewidth]{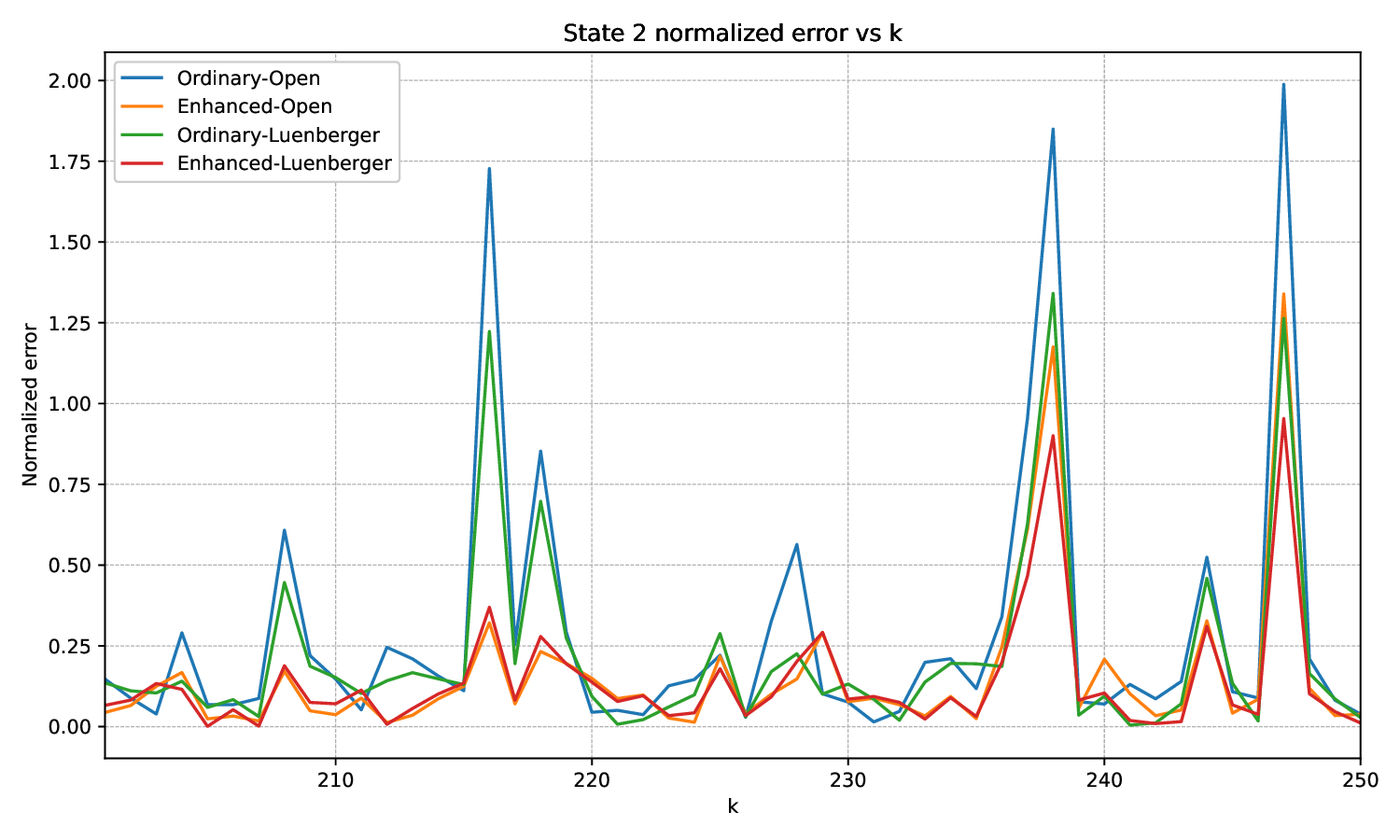}
    \caption{The second component of normalized state error vs $k$ after stable.}
    \label{stablee2}
  \end{subfigure}
  \caption{
  A visual comparison illustrating the validity of the proposed learning refinement. The curves labeled \emph{Real} denote the ground truth trajectory, while \emph{Ordinary-Open} and \emph{Ordinary-Luenberger} correspond to nominal open-loop and Luenberger observers. \emph{Enhanced-Open} and \emph{Enhanced-Luenberger} show the respective observers after learning-based parameter refinement.}
  \label{Visual}
\end{figure}

A visual example is shown in Fig.~\ref{Visual}. The true system matrices and initial states are

\begin{equation}
\begin{aligned}
&A_{real}=\begin{pmatrix}
1.0200&0.6800\\
-0.6800&0.3400
\end{pmatrix},\ 
B_{real}=\begin{pmatrix}
1.5000\\
0.7000
\end{pmatrix},
\\
&C_{real}=\begin{pmatrix}
1.0000 & 0.0000
\end{pmatrix},\ 
x_{0,real}=\begin{pmatrix}
0.4617\\
0.2674
\end{pmatrix}
\end{aligned}
\end{equation}
while the nominal parameters used for initialization are
\begin{equation}
\begin{aligned}
&A_{init}:=\begin{pmatrix}
1.0368&0.6864\\
-0.6683&0.3515
\end{pmatrix},\ 
B_{init}:=\begin{pmatrix}
1.4439\\
0.6907
\end{pmatrix},
\\
&C_{init}:=\begin{pmatrix}
1.1104 & -0.0319
\end{pmatrix},\ 
x_{0,init}=\begin{pmatrix}
5.8107\\
8.3609
\end{pmatrix}
\end{aligned}
\end{equation}

As the figure illustrates, the LEO significantly reduces the normalized estimation error for the steady state. Both the open-loop and Luenberger configurations benefit from the proposed refinement, demonstrating improved accuracy and robustness to model perturbations.

\section{Conclusion}

This paper presented a learning-enhanced framework for designing observers for linear time-invariant systems under modest parameter uncertainty. Rather than relying solely on nominal models, the proposed method refines the system matrices through a gradient-based optimization of a data-driven loss that captures steady-state output discrepancies. The optimized model then yields an improved observer with substantially reduced estimation error. Extensive Monte Carlo studies across various system dimensions confirm consistent performance gains for both open-loop and closed-loop observers, with average normalized error reductions exceeding 15\%, and statistical tests also verify the significance of the improvement.

Overall, the LEO framework provides a systematic and computationally scalable means to integrate learning-based parameter refinement into classical observer design. The results indicate that learning-assisted observer refinement offers a promising avenue for enhancing estimation accuracy in uncertain dynamical environments. Future directions include establishing theoretical convergence guarantees, analyzing robustness under structured or adversarial disturbances, and extending the approach to time-varying or nonlinear settings.

\printbibliography

@ARTICLE{Luenberger1964Observing,
  author={Luenberger, David G.},
  journal={IEEE Transactions on Military Electronics}, 
  title={Observing the State of a Linear System}, 
  year={1964},
  volume={8},
  number={2},
  pages={74-80},
  doi={10.1109/TME.1964.4323124}
}

@article{Kalman1960A,
    author = {Kalman, R. E.},
    title = {A New Approach to Linear Filtering and Prediction Problems},
    journal = {Journal of Basic Engineering},
    volume = {82},
    number = {1},
    pages = {35-45},
    year = {1960},
    month = {03},
    issn = {0021-9223},
    doi = {10.1115/1.3662552},
    url = {https://doi.org/10.1115/1.3662552}
}

@ARTICLE{Narendra1976Stable,
  author={Narendra, K.S. and Valavani, L.S.},
  journal={Proceedings of the IEEE}, 
  title={Stable adaptive observers and controllers}, 
  year={1976},
  volume={64},
  number={8},
  pages={1198-1208}
  }

@ARTICLE{Kreisselmeier1977Adaptive,
  author={Kreisselmeier, G.},
  journal={IEEE Transactions on Automatic Control}, 
  title={Adaptive observers with exponential rate of convergence}, 
  year={1977},
  volume={22},
  number={1},
  pages={2-8}
}

@ARTICLE{Nuyan1979Minimal,
  author={Nuyan, S. and Carroll, R.},
  journal={IEEE Transactions on Automatic Control}, 
  title={Minimal order arbitrarily fast adaptive observers and identifiers}, 
  year={1979},
  volume={24},
  number={2},
  pages={289-297}
}

@ARTICLE{Katiyar2023Initial,
  author={Katiyar, Atul and Roy, Sayan Basu and Bhasin, Shubhendu},
  journal={IEEE Transactions on Automatic Control}, 
  title={Initial-Excitation-Based Robust Adaptive Observer for MIMO LTI Systems}, 
  year={2023},
  volume={68},
  number={4},
  pages={2536-2543}
  }

@ARTICLE{Romero2025An,
  author={Romero, Jose Guadalupe and Ortega, Romeo and Ushirobira, Rosane and Efimov, Denis},
  journal={IEEE Transactions on Automatic Control}, 
  title={An Adaptive State Observer for State-Affine Systems with Uncertain Parameters and Unknown Additive Output Disturbances}, 
  year={2025},
  volume={},
  number={},
  pages={1-6}
  }

@ARTICLE{Carroll1973An,
  author={Carroll, R. and Lindorff, D.},
  journal={IEEE Transactions on Automatic Control}, 
  title={An adaptive observer for single-input single-output linear systems}, 
  year={1973},
  volume={18},
  number={5},
  pages={428-435},
  doi={10.1109/TAC.1973.1100367}
}

@ARTICLE{Nagumo1967A,
  author={Nagumo, J. and Noda, A.},
  journal={IEEE Transactions on Automatic Control}, 
  title={A learning method for system identification}, 
  year={1967},
  volume={12},
  number={3},
  pages={282-287}
}

@ARTICLE{Farison1967Identification,
  author={Farison, J. and Graham, R. and Shelton, R.},
  journal={IEEE Transactions on Automatic Control}, 
  title={Identification and control of linear discrete systems}, 
  year={1967},
  volume={12},
  number={4},
  pages={438-442}
}

@article{Chu1994Simultaneous,
title = {Simultaneous Parameter Identification and States Estimation using Neural Networks},
journal = {IFAC Proceedings Volumes},
volume = {27},
number = {8},
pages = {633-637},
year = {1994},
note = {IFAC Symposium on System Identification (SYSID'94), Copenhagen, Denmark, 4-6 July},
issn = {1474-6670},
doi = {https://doi.org/10.1016/S1474-6670(17)47780-9},
url = {https://www.sciencedirect.com/science/article/pii/S1474667017477809},
author = {Chu, S. Reynold and Shoureshi, Rahmat}
}

@INPROCEEDINGS{Mohamed2023The,
  author={Mohamed, Mohamed Naveed Gul and Goyal, Raman and Chakravorty, Suman and Wang, Ran},
  booktitle={2023 American Control Conference (ACC)}, 
  title={The Information-State Based Approach to Linear System Identification}, 
  year={2023},
  volume={},
  number={},
  pages={301-306}
}

@article{Bernstein1989Steady,
title = {Steady-state Kalman filtering with an H$\infty$ error bound},
journal = {Systems $\&$ Control Letters},
volume = {12},
number = {1},
pages = {9-16},
year = {1989},
issn = {0167-6911},
doi = {https://doi.org/10.1016/0167-6911(89)90089-3},
url = {https://www.sciencedirect.com/science/article/pii/0167691189900893},
author = {Bernstein, Dennis S. and Haddad, Wassim M.}
}

@INPROCEEDINGS{Douglas1991Process,
  author={Douglas, Randal K. and Speyer, Jason L.},
  booktitle={1991 American Control Conference}, 
  title={Process and Sensor Noise Robustness in Detection Filter Design}, 
  year={1991},
  volume={},
  number={},
  pages={2274-2279},
  doi={10.23919/ACC.1991.4791807}
}

@INPROCEEDINGS{Edelmayer1994An,
  author={Edelmayer, A. and Bokor, J. and Keviczky, L.},
  booktitle={Proceedings of 1994 33rd IEEE Conference on Decision and Control}, 
  title={An $H_{\infty}$ filtering approach to robust detection of failures in dynamical systems}, 
  year={1994},
  volume={3},
  number={},
  pages={3037-3039 vol.3},
  doi={10.1109/CDC.1994.411321}
}

@article{Edelmayer1996H,
title = {$H_{\infty}$ Detection Filter Design for Linear Systems: Comparison of Two Approaches},
journal = {IFAC Proceedings Volumes},
volume = {29},
number = {1},
pages = {6359-6364},
year = {1996},
note = {13th World Congress of IFAC, 1996, San Francisco USA, 30 June - 5 July},
issn = {1474-6670},
doi = {https://doi.org/10.1016/S1474-6670(17)58701-7},
url = {https://www.sciencedirect.com/science/article/pii/S1474667017587017},
author = {Edelmayer, A. and Bokor, J. and Keviczky, L.}
}

\end{document}